\documentclass{styles/svproc}

\usepackage{url}

\usepackage[numbers]{natbib}
\usepackage{graphicx}
\usepackage{textcomp}
\usepackage{xcolor}
\usepackage{float}
\usepackage{caption}
\usepackage{subcaption}
\usepackage{amsmath}
\usepackage{amssymb}
\usepackage{authblk}
\usepackage[english]{babel}

\begin{document}
\mainmatter              
\title{Self-Awareness In Intelligent Vehicles: Experience based Abnormality Detection}
\titlerunning{Self-Awareness in IV: Experience Based Abnormality Detection}

\author{Divya Kanapram \textsuperscript{1,3}, Pablo Marin-Plaza \textsuperscript{2},  Lucio Marcenaro \textsuperscript{1}, David Martin \textsuperscript{2} Arturo de la Escalera \textsuperscript{2}, Carlo Regazzoni\textsuperscript{1}}

\institute{\textsuperscript{1}University of Genova, Italy. divya.kanapram@ginevra.dibe.unige.it, \{carlo.regazzoni,lucio.marcenaro\}@unige.it\\ %
\textsuperscript{2}Universidad Carlos III, Leganes Spain. \{pamarinp, dmgomez, escalera\}@ing.uc3m.es\\ %
\textsuperscript{3}Queen Mary University of London, UK.\\%
}

\authorrunning{Divya et al.}

\maketitle              
\begin{abstract}

The evolution of Intelligent Transportation System in recent times necessitates the development of self-driving agents: the self-awareness consciousness. This paper aims to introduce a novel method to detect abnormalities based on internal cross-correlation parameters of the vehicle. Before the implementation of Machine Learning, the detection of abnormalities were manually programmed by checking every variable and creating huge nested conditions that are very difficult to track. Nowadays, it is possible to train a Dynamic Bayesian Network (DBN) model to automatically evaluate and detect when the vehicle is potentially misbehaving. In this paper, different scenarios have been set in order to train and test a switching DBN for Perimeter Monitoring Task using a semantic segmentation for the DBN model and Hellinger Distance metric for abnormality measurements.


\end{abstract}

\section{Introduction}

\label{section I}
The self-awareness field is vast in terms of detecting abnormalities in the field of Intelligent Vehicles \cite{xiong2010autonomous}. It is possible to classify critical, medium, or minor abnormalities by defining the line between normal and abnormal behaviour with the help of top design architectures. The problem of self-awareness systems is to measure every sensor, data acquired, and behavior of the system at every moment, comparing each measurement with the nominal range. Due to the huge amount of data, these tasks are not easy and becomes typically dead-end in big projects where the re-usability is not possible. Furthermore, it is possible to be unaware of situations where the vehicle is not working in the normal ranges for a very short period of time. Self-awareness management could be divided into three main categories which are hardware, software, and behavior. The first category is based on the detection of malfunctions on electronic devices, actuators, sensors, CPUs, communication, etc. The second category focuses on software requirements where the most important measurements for message delivery are time, load, bottlenecks, delays, heartbeat, among others. Finally, the last self-awareness field analyzes the behavior of the vehicle which is related to the performance of the task assigned at each moment such as keep in lane, lane change, intersection management, roundabout management, overtaking, stop, etc. Accordingly, the management of self-awareness is a cross-layer problem where every manager should be built subsequently to the other layers to create a coherent self-awareness system \cite{schlatow2017self}.
 
To reduce the amount of process effort in intelligent self-awareness system, the emergent techniques in Machine Learning allow the creation of models using Dynamic Bayesian Networks (DBN) to automatize this process~\cite{lewis2016self}. The novelty of this paper is the use of DBN models to generate a cross-correlation between a pair of internal features of the vehicle using a Hellinger distance metric for abnormality detection. Finally, compared the performance of different DBN models in order to select the best model for abnormality detection.

The remainder of this paper is organized as follows. Section 2 presents a survey of the related work. In section 3, described the proposed method, defining principles exploited in the training phase and the steps involved in the test phase for detecting the abnormality. Section 4 summarizes the experimental setup in addition to the description of the research platform used. Section 5 gathered the results (i.e., abnormality measurements) from pair based DBNs for each vehicle and compared the results, and finally, section 6 concludes the paper.
\setlength{\parskip}{-10pt}
\section{State of the art}
\setlength{\parskip}{-10pt}
This section describes some of the related work regarding the development of self-awareness in \textit{agents}. In \cite{baydoun2018multi}, the authors propose an approach to develop a multilevel self-awareness model learned from multisensory data of an agent. Such a learned model allows the agent to detect abnormal situations present in the surrounding environment. In another work \cite{ravanbakhsh2018learning}, the learning of the self-awareness model for autonomous vehicles based on the data collected from human driving. On the other hand, in \cite{leite2018safety}, the authors propose a new architecture for mobile robots with a model for risk assessment and decision making when detecting difficulties in the autonomous robot design. In \cite{xie2018driving}, the authors proposed a model of driving behavior awareness (DBA) that can infer driving behaviours such as lane change. In all the above works either used the data from one entity or the objective was limited; for example in \cite{xie2018driving} the objective was to detect lane change either on left or right side of the considered vehicles. In this work, we have considered the data from the real vehicles and developed pair based switching DBN models for each vehicle and finally made the performance comparison among different DBNs learned.
\setlength{\parskip}{-12pt}
\section{Proposed method}
\label{section II}
This section discusses how to develop “intelligence” and “awareness” into vehicles to generate “Self-aware intelligent vehicles.” The first step is to perform synchronization operation over the acquired multisensory data to make them synchronized in time in a way to match their time stamps. The data sets collected for training and testing are heterogeneous, and two vehicles are involved in the considered scenarios. The observed multisensory data from the vehicles are partitioned into different groups to learn a pair-based switching DBN model for each pair-based vehicle feature. Then compare the performances to qualify the best pair-based feature for automatic detection of abnormality. Switching DBNs are probabilistic models to integrate observations received from multiple sensors in order to understand the environment where they operate and take appropriate actions in different situations. The proposed method is divided into two phases: offline training and online testing. In the offline training phase, learn DBNs from the experiences of the vehicle in their normal behaviour. In the next phase, online testing, we have used a dynamic data series that are collected from the vehicles while they pass through different experiences than in the training phase. Accordingly, a filter called Markov Jump Particle Filter (MJPF) applied to the learned DBN models to estimate the future states of the vehicles and finally detects the abnormality situations present in the environment.
\subsection{Offline training phase}
\vspace*{6px}
In this phase, learn switching DBNs from the datasets collected from the experiences of the vehicle in their normal behaviour. The various steps involved in learning a DBN model are described below. 
\vspace{-2mm}
\subsubsection{Generalized states}
The intelligent vehicles used in this work are equipped with one lidar of 16 layers and 360 degrees of Field of view(FOV), a stereo camera, and encoder devices to monitor different tasks being performed. In this work, it is assumed that each vehicle is aware of the other vehicle by its communication scheme and cooperation skills. By considering the vehicles endowed with a certain amount of sensors that monitors its activity, it is possible to define $Z^c_k$ as any combination (indexed by $c$) of the available measurements in a time instant $k$. Let $X^c_k$ be the states related to measurements $Z^c_k$, such that: $Z^c_k = X^c_k + \omega_k$; where $\omega_k$ represents the sensor noise. The generalized states of a sensory data combination $c$ can be defined as: 

\vspace{1mm}
\begin{equation}\label{eq1}
\boldsymbol{X}_k^c = [X_k^c \hspace{2mm}  \dot{X}_k^c \hspace{2mm} \ddot{X}_k^c  \hspace{0.5mm} \cdots \hspace{0.5mm} X^{c,(L)}_k]^\intercal,
\end{equation}
where $(L)$ indicates the $L$-th time derivative of the state.
\subsubsection {Vocabulary generation and state transition matrix calculation}
In order to learn the desecrate level of the DBN (i.e., the orange outlined box in Fig. \ref{fig:DBN}), it is required to map the generalized states into a set of nodes. We have used a clustering algorithm called Growing Neural Gas (GNG) to group these generalized states and to obtain nodes. In GNG, the learning is continuous, and the addition or removal of nodes is automated \cite{fritzke1995growing}. These are the main reasons to choose GNG algorithm over other clustering algorithms such as K-means \cite{hartigan1979algorithm} or self-organizing map (SOM) \cite{kohonen1990self}. The output of each GNG consists of a set of nodes that encode constant behaviours for specific sensory data combinations time derivative order. In other words, at a time instant, each GNG takes the data related to a single time derivative $X_k^{c,(i)} \in \boldsymbol{X}_k^c $ and cluster it with the same time derivative data acquired in previous time instances. The nodes associated with each GNG can be called as a set of \textit{letters} containing the main behaviours of generalized states. The collection of nodes encoding $i$-th order derivatives in an observed data combination $c$ is defined as follows:
\vspace{-2mm}
\begin{equation}\label{eq3}
A^c_i = \{\bar{X}^{(i),c}_{1}, \bar{X}^{(i),c}_{2}, \cdots  \bar{X}^{(i),c}_{P^c_i}\}
\end{equation}
where $P^c_i$ is the number of nodes in the GNG associated to the $i$-th order derivative of data in data combination $c$. Each node in a GNG represents the centroids of associated data point clusters. By taken into consideration all the possible combinations of centroids obtained from GNGs, we can get a set of \textit{words}, that define generalized states in an entirely semantic way. The obtained \textit{words} can form a \textit{dictionary} and can be defined as:\\
\vspace*{-\baselineskip}
\begin{equation}\label{eq4}
D^c = \{\beta, \dot{\beta}, \cdots  \beta^{(L)}\}
\end{equation}
\vspace{-1mm}
where $\beta^{(i)}_c \in A^c_i$. $D^c$ contains all possible combinations of discrete generalized states. $D^c$ is a high-level hierarchy variable that explains the system states from a semantic viewpoint. In this work, we have only considered states, and it's first-order derivatives.\\ Furthermore, estimated the state transition matrices based on the timely evolution of such \textit{letters} and \textit{words}. The state transition matrix is a matrix that contains the transition probability between the discrete level nodes of the switching DBN shown in Fig. \ref{fig:DBN}. When the first data emission occurs, the state transition matrix provides the probability of the next discrete level, i.e., the probability of activation of a node from the GNG associated with the first-order derivatives of the states. The vocabulary (i.e.,\textit{letters}, \textit{words} and \textit{dictionary} and the transition matrices constitute the discrete level of the DBN model. 
\subsubsection{DBN model Learning}
\begin{figure}[h]
\centering
 	\includegraphics[width = 0.7 \linewidth ]{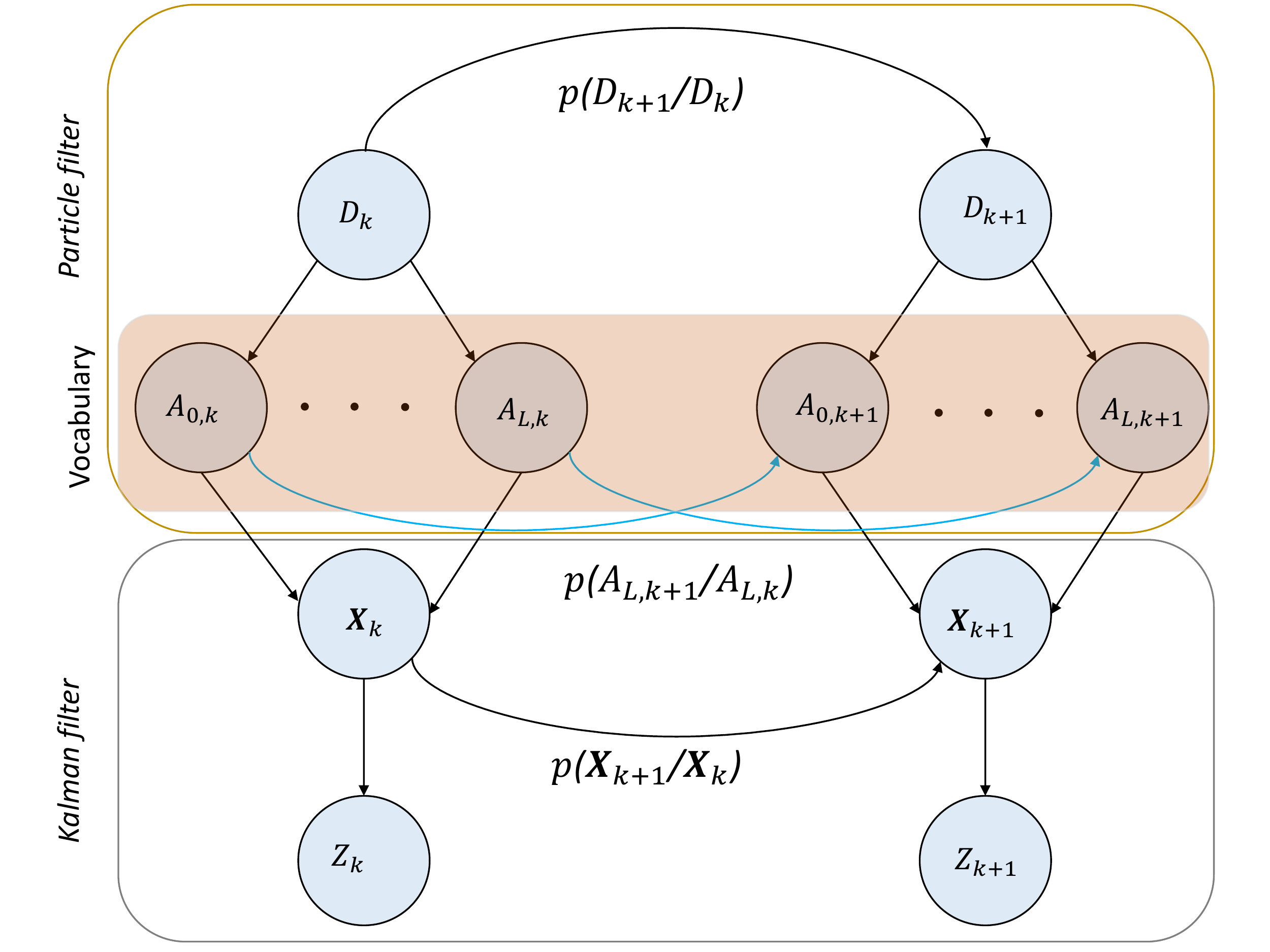}
\caption{Proposed DBN}
\label{fig:DBN}
\end{figure}

All the previous steps are the step by step learning process of the switching DBNs by each entity taken into consideration. The number of DBNs learned by each entity in the network can be written as:
\vspace{-3mm}
\begin{equation}\label{eq2}
 \boldsymbol{DBN}m = \{DBN_1, \cdots, DBN_n\}.
\end{equation}
\vspace{-1mm}
where $m$ represents the $m^{\text{th}}$ vehicle in the network and $n$ is the total number DBN learned by the $m^{\text{th}}$ vehicle. The same DBN architecture is considered for making inferences with different sensory data combinations belong to different vehicles. The learned DBN can be represented as shown in the Fig.\ref{fig:DBN}. The DBN has mainly three levels such as measurements, continuous and discrete levels.

\subsection{Online test phase}
\vspace*{4px}
In this phase, we have proposed to apply a dynamic switching model called Markov Jump Particle Filter (MJPF) \cite{baydoun2018learning} to make inferences on the learned DBN models.  MJPF is a mixed approach with particles inside each Kalman filter. The MJPF is able to predict and estimate continuous and discrete future states and to detect deviations from the normal model. In MJPF, we use Kalman Filter (KF) in state-space and Particle Filter (PF) in super state space in Fig. \ref{fig:DBN}.
\vspace{-1mm}
\subsubsection{Abnormality detection and complementarity check}
In probability theory, a statistical distance quantifies the distance between two statistical objects, which can be two random variables, or two probability distributions, etc. Some important statistical distances include: Bhattacharya distance \cite{bhattacharyya1943measure}, Hellinger Distance \cite{pardo2005statistical}, Jensen–Shannon divergence \cite{endres2003new}, Kullback–Leibler (KL) divergence \cite{hershey2007approximating} etc. and they are the distances generally used between two distributions.
Although, the HD is defined between vectors having only positive or zero elements \cite{abdi2007encyclopedia}. The datasets in this work are normalized, so the values vary between zero and one; there aren’t any negative values. Moreover, HD is symmetric compared to KL divergence. By these reasons, HD is more appropriate than using other distance metrics as abnormality measure. Moreover, the works in \cite{Lourenzutti2014} and \cite{baydoun2019prediction} used HD as an abnormality measurement.\\
Abnormality measurement can be defined as the distance between predicted state values and the observed evidence. Accordingly, let $p(\boldsymbol{X}^c_k|\boldsymbol{X}^c_{k-1})$ be the predicted generalized states and $p(Z_k|\boldsymbol{X}^m_k)$ be the observation evidence. The HD can be written as:
\begin{equation}\label{eq6}
\theta_k^c = \sqrt{1 - \lambda_k^c},
\end{equation}
where $\lambda_k^c$ is defined as the Bhattacharyya coefficient \cite{Bhattacharyya1943}, such that:
\begin{equation}\label{eq7}
\lambda^c_k = \int \sqrt{p(\boldsymbol{X}_k^c|\boldsymbol{X}_{k-1}^c) p(Z_k^c|\boldsymbol{X}_k^c)} \hspace{1mm} \mathrm{d}\boldsymbol{X}_k^c.
\end{equation}
\vspace{-1mm}
When a given experience in evaluation an abnormality measurement obtains at each time instant, and can be seen in the equation \eqref{eq6}. The variable $\theta_k^c \in [0,1]$, where values close to $0$ indicate that measurements match with predictions; whereas values close to $1$ reveal the presence of an abnormality. After calculating the abnormality measures by HD, it is possible to check the complementarity among different DBN models learned.
\vspace{-1mm}
\section{Experimental Setup}
\begin{figure}[h]
	\begin{subfigure}[t]{0.5\textwidth}
		\centering
		\includegraphics[width=4cm,height=3cm]{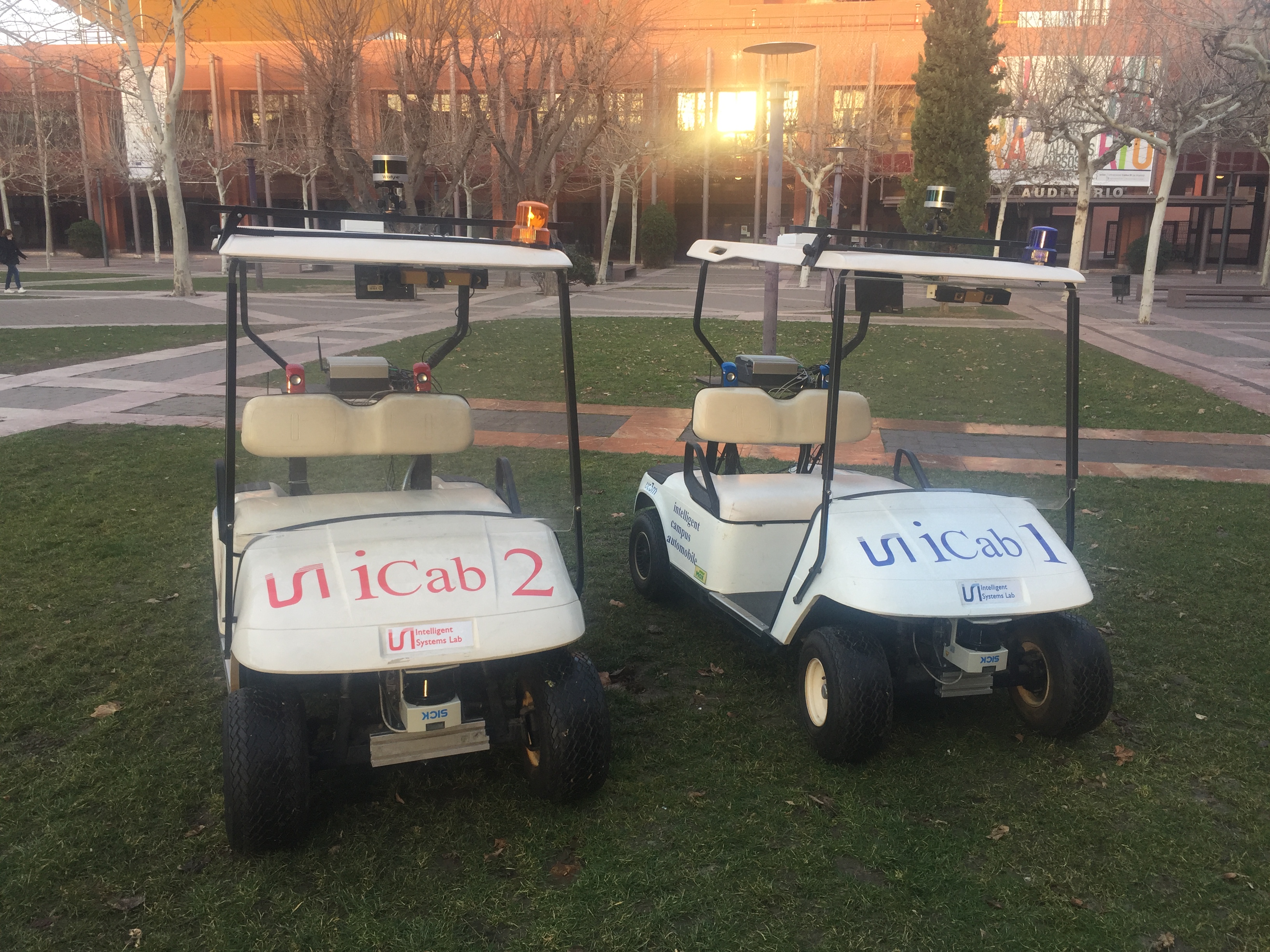}
		\caption{The autonomous vehicles (iCab)}
		\label{fig:iCab}
	\end{subfigure}%
	\begin{subfigure}[t]{0.5\textwidth}
		\centering
		\includegraphics[width=4cm,height=3cm]{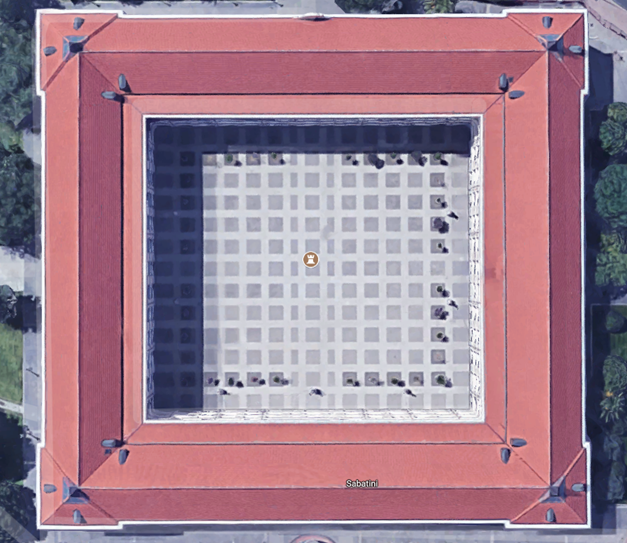}
		\caption{The environment}
		\label{fig:environment}
	\end{subfigure}
	\caption{The agents and the environment used for the experiments.}
	\label{fig:icabAndEnv}
\end{figure}

\vspace{2mm}
\label{section III}
In order to validate the proposed method, it has been used two intelligent research platform called iCab (\textbf{I}ntelligent \textbf{C}ampus \textbf{A}utomo\textbf{B}ile)\cite{marin2018global} (see Fig.\ref{fig:iCab}) with autonomous capabilities. To process and navigate through the environment, the vehicles count with two powerful computers along with the screen for debugging and Human-Machine Interaction (HMI) purposes. The software prototyping tool used is ROS \cite{quigley2009ros}.

The data sets collected with the two \textit{iCab} vehicles are synchronized in order to observe the vehicles as different entities in a heterogeneous way to match their time stamps.
The intercommunication scheme is proposed in \cite{kokuti2017v2x} where both vehicles share all its information over the network by a Virtual Private Network(VPN). For this experiment, as long as the synchronization level reaches the nanoseconds, the recorded dataset in both vehicles has been merged and ordered using the timestamp generated by the clock on each vehicle which has been previously configured with a Network Time Protocol (NTP) tool called \textit{Chrony}. Both vehicles perform a PMT task which consists of the autonomous movement of platooning around a square building (see Fig.\ref{fig:environment}). The data generated from the lidar odometry such as the ego-motion of the vehicle and the different combinations of the control variables such as steering angle, rotor velocity and power of the rotor are considered the main metrics to learn and test the models. Moreover, it aims to detect the unseen dynamics of the vehicles with the proposed method.
\begin{figure}[h]
	\begin{subfigure}[t]{0.5\textwidth}
		\centering
		\includegraphics[height=4.5cm, width=5.5cm]{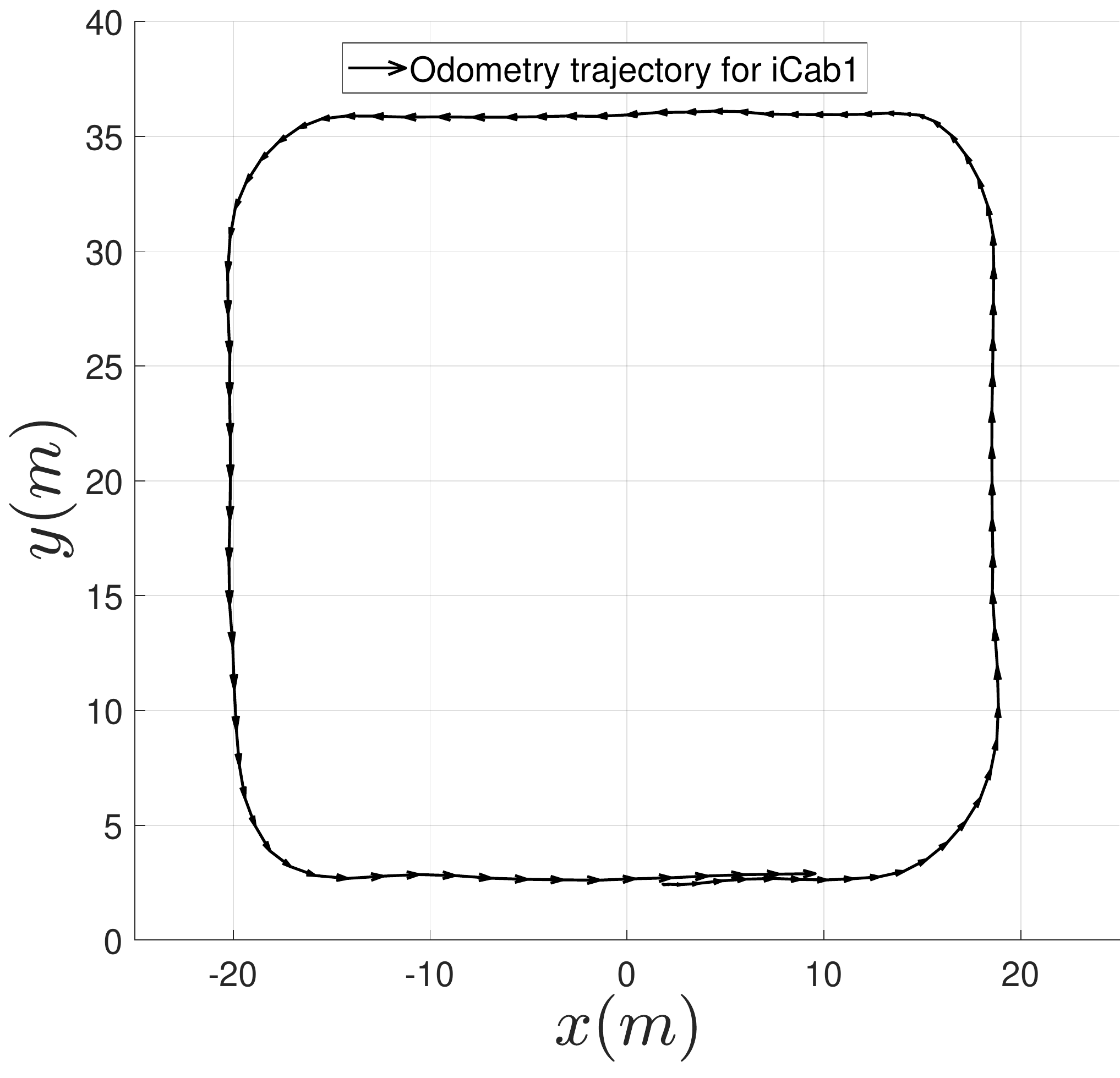}
		\caption{Perimeter monitoring}
		\label{fig:PM1}
	\end{subfigure}%
	\begin{subfigure}[t]{0.5\textwidth}
		\centering
		\includegraphics[height=4.5cm, width=5.5cm]{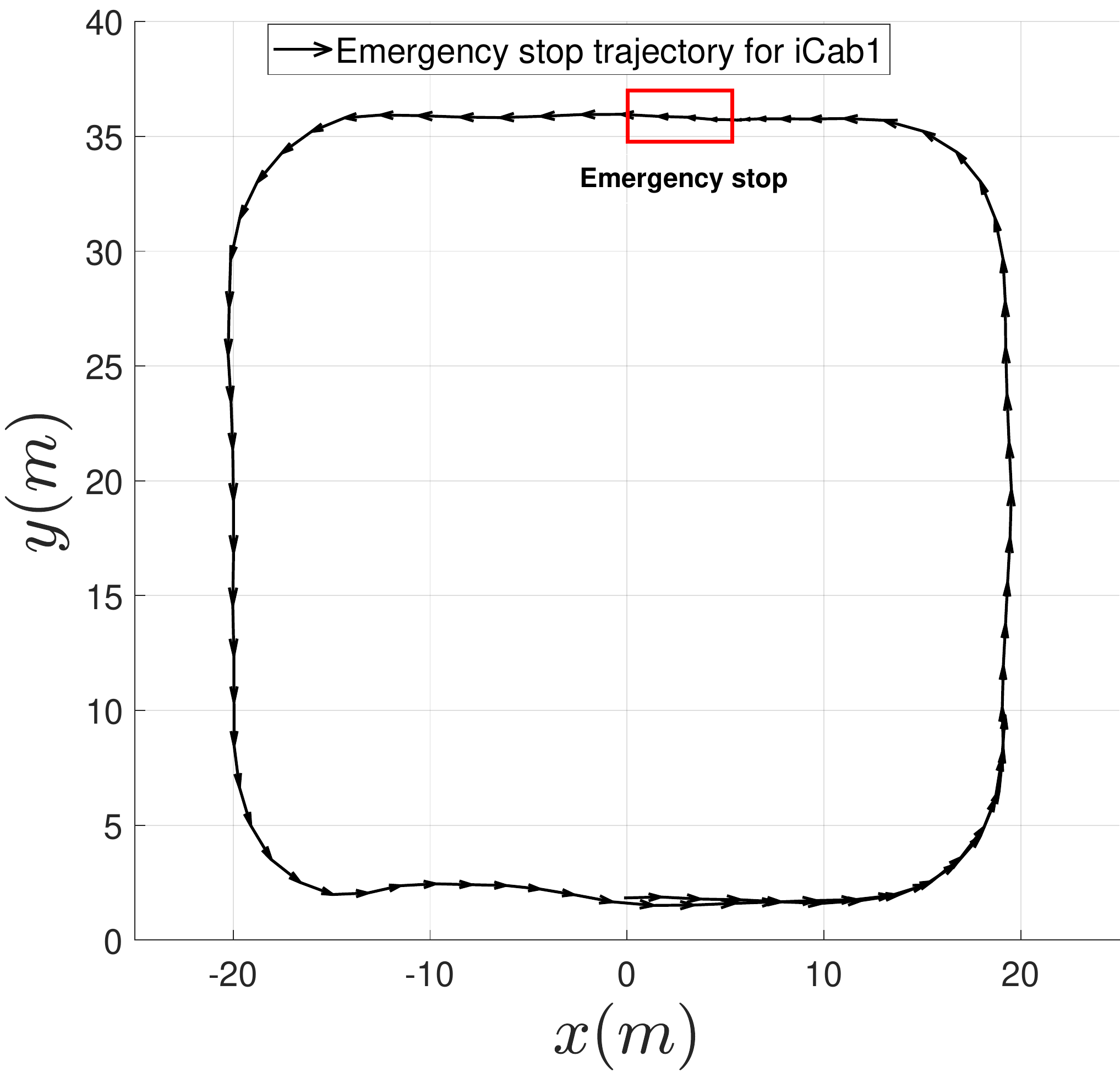}
		\caption{Emergency stop}
		\label{fig:ES1}
	\end{subfigure}
	\caption{Odometry data for iCab1}
	\label{fig:Position data iCab1}
\end{figure}
\begin{figure}[h]
	\begin{subfigure}[t]{0.5\textwidth}
		\centering
		\includegraphics[height=4.5cm, width=5.5cm]{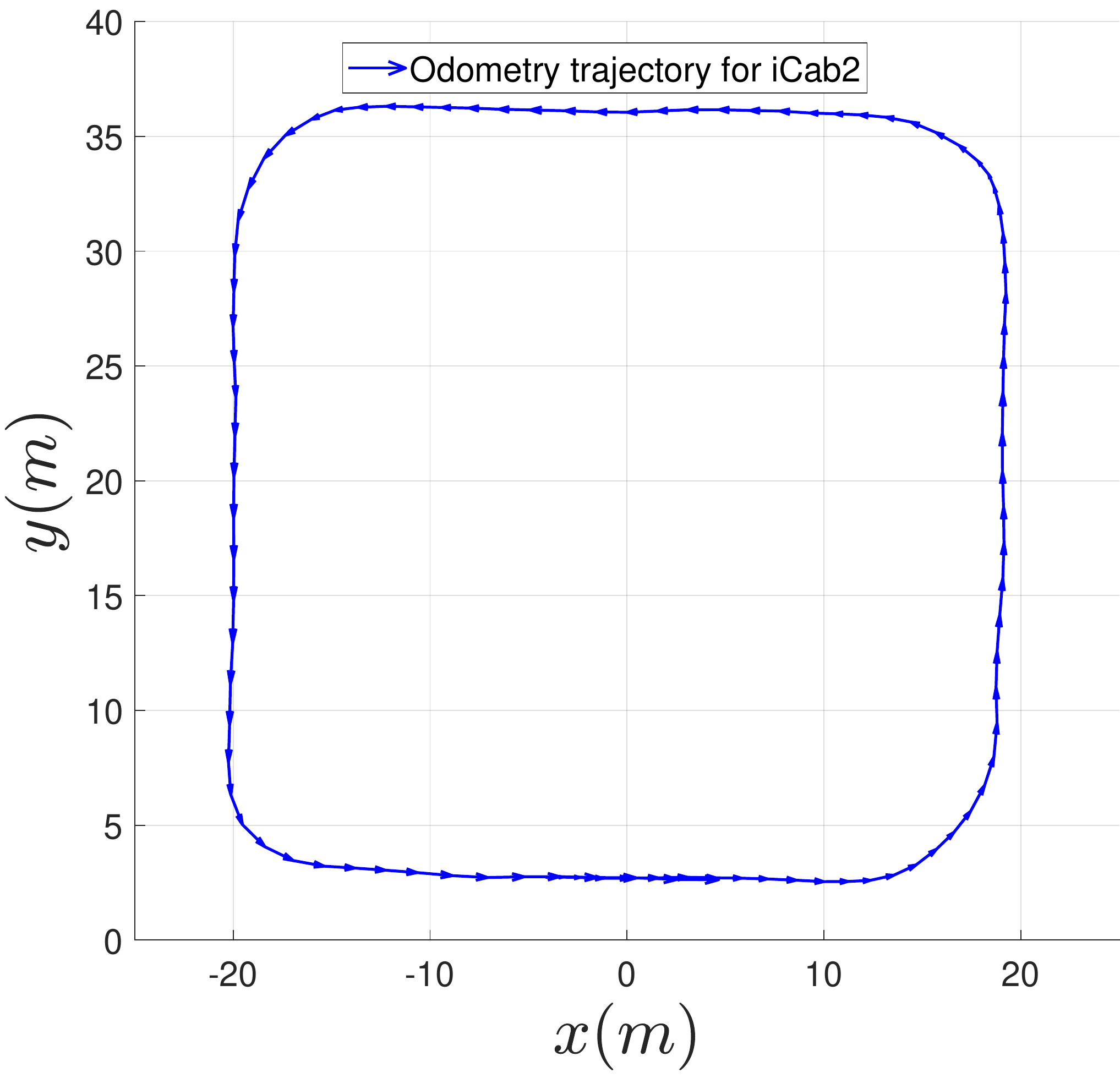}
		\caption{Perimeter monitoring}
		\label{fig:PM2}
	\end{subfigure}%
	\begin{subfigure}[t]{0.5\textwidth}
		\centering
		\includegraphics[height=4.5cm, width=5.5cm]{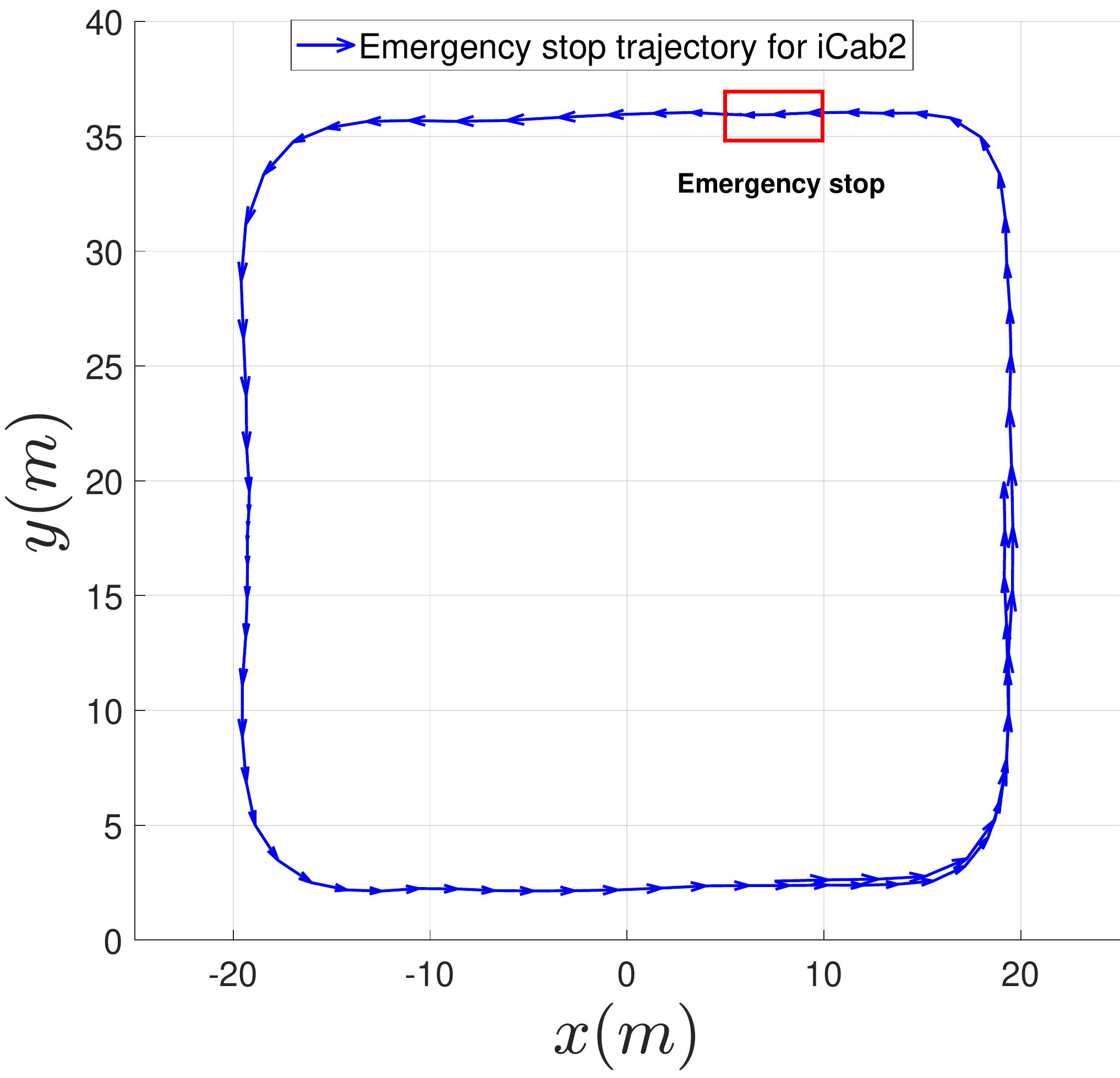}
		\caption{Emergency stop}
		\label{fig:ES2}
	\end{subfigure}
	\caption{Odometry data for iCab2}
	\label{fig:Position data Icab2}
\end{figure}
\begin{figure}[h]
	\begin{subfigure}[t]{0.5\textwidth}
		\centering
		\includegraphics[height=5cm, width=6cm]{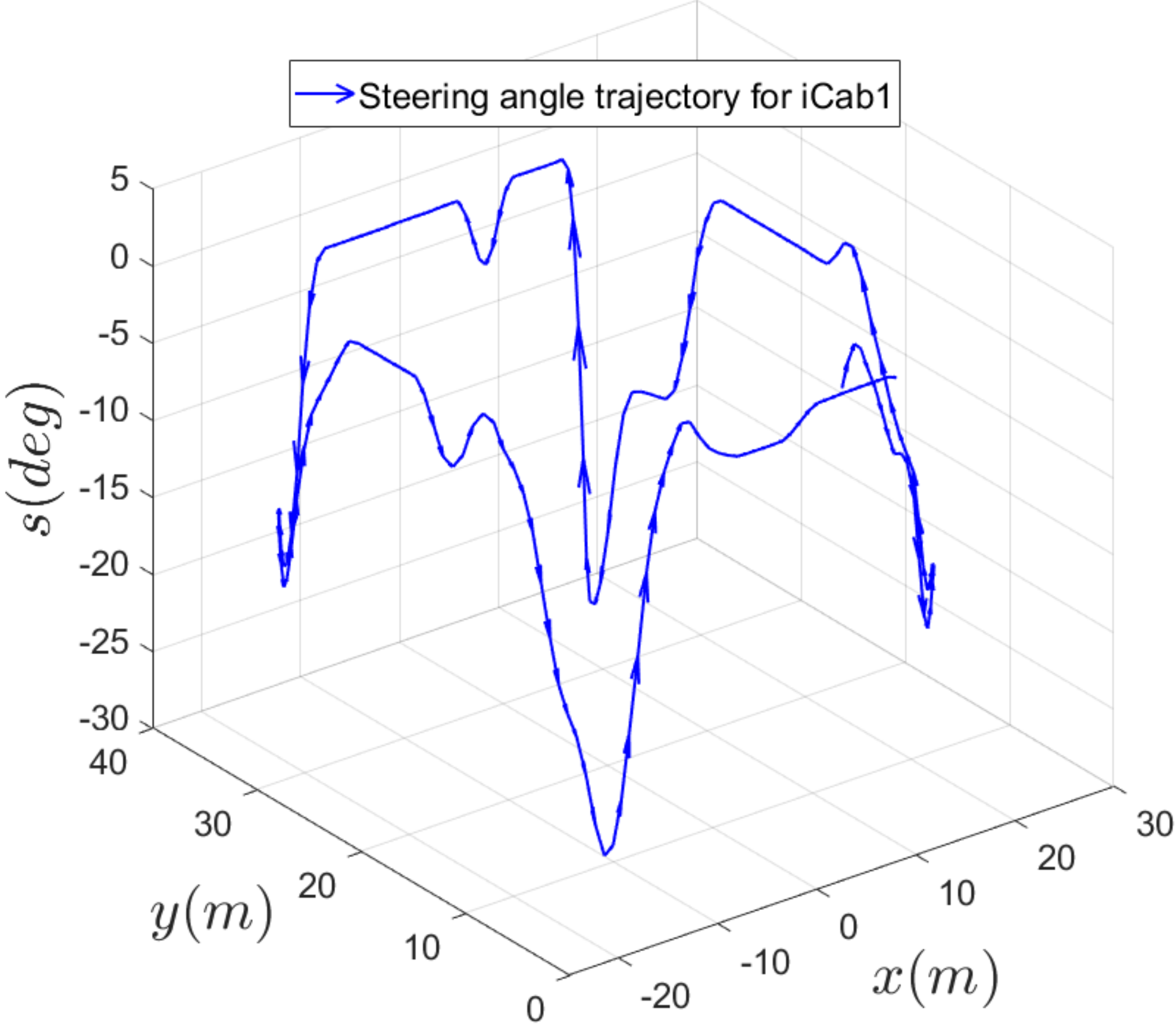}
		\caption{Steering angle(\textit{s}) w.r.t position}
		\label{fig:S}
	\end{subfigure}%
	\begin{subfigure}[t]{0.5\textwidth}
		\centering
		\includegraphics[height=5cm, width=6cm]{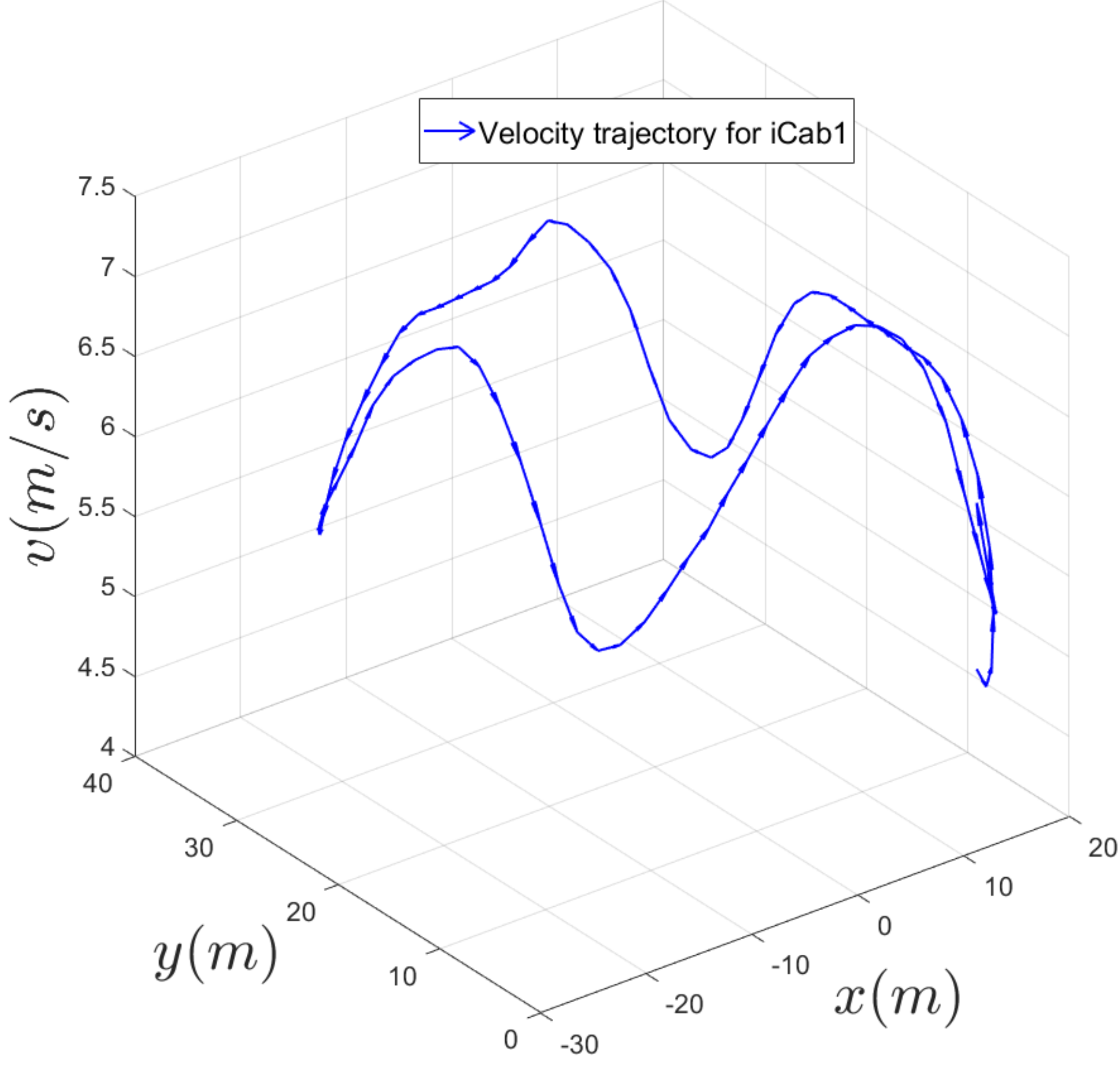}
		\caption{Velocity(\textit{v}) w.r.t position}
		\label{fig:V}
	\end{subfigure}
	\caption{Control data for iCab1 for perimeter monitoring task}
	\label{fig:Control data}
\end{figure}
\begin{figure}[h]
\centering
 	\includegraphics[height=5cm, width=6cm ]{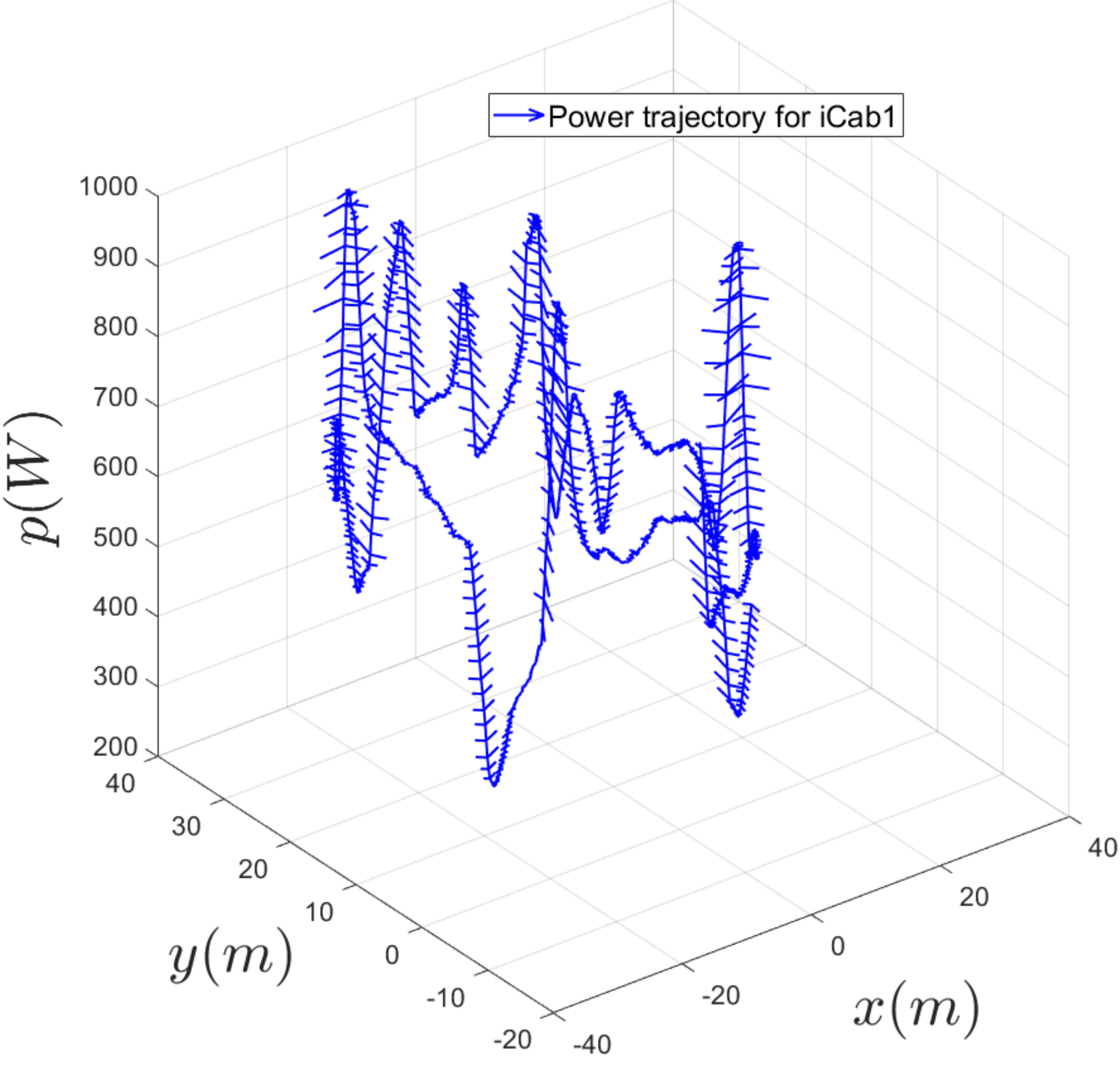}
\caption{Rotor power(\textit{p}) w.r.t position for iCab1 for perimeter monitoring task}
\label{fig:P}
\end{figure}
\setlength{\parskip}{-12pt}
\subsection{Perimeter Monitoring Task (PMT)}
\vspace{0.5\baselineskip}
In order to generate the required data for learning and detect abnormalities, both vehicles perform a rectangular trajectory in a platooning mode. The leader just follows the rectangular path and the follower receives the path and keep the desired distance with the leader. This task has been divided into two different scenarios. 
\begin{itemize}
    \item \textbf{Scenario I}: Both vehicles perform the platooning operation by following a rectangular trajectory in a closed environment, as shown in Fig. \ref{fig:environment}, four laps in total by recording the ego-motion, stereo camera images, lidar Point Cloud Data, encoders, and self-state. Notice that the GPS has troubles to acquire good signal because of the urban canyon.
    Fig.\ref{fig:PM1} and Fig.\ref{fig:PM2} show the plots of odometry data for the perimeter monitoring task for iCab1 and iCab2 respectively. Moreover, Fig.\ref{fig:Control data} shows the steering angle w.r.t the iCab1's position(Fig.\ref{fig:S}) and rotor velocity w.r.t the iCab1's position(Fig.\ref{fig:V}). The rotor power data plotted w.r.t iCab1's position is shown in Fig.\ref{fig:P}.
    \item \textbf{Scenario II}: Both vehicles perform the same experiment, but now a pedestrian crosses in front of the leader vehicle(i.e., iCab1). When the leader vehicle detects the pedestrian, automatically executes a stop and wait until the pedestrian fully crosses and move out from the danger zone. Meanwhile, the follower (i.e., iCab2) detects the stop of the leader and stops at a certain distance. When the leader continues the PMT, the follower mimics the action of the leader. Fig.\ref{fig:ES1} and Fig.\ref{fig:ES2} show the plots of odometry data for the emergency stop criteria for iCab1 and iCab2 respectively.
    
\end{itemize}
\setlength{\parskip}{-10pt}
\section{Results}

\label{section IV}
As explained in the previous section, there are two different scenarios taken into consideration with two vehicles. Moreover, the data combinations from odometry and control of vehicles have been treated independently and finally compared the abnormality measures to understand the correlation between them. We set the abnormality threshold to 0.4, considering the average Hellinger distance value of 0.2 when vehicles operate in normal conditions. The DBN models are trained over the scenario I based on PMT where no pedestrians are crossing in front of the vehicles. As said at the beginning of this paper, one of the objectives is the automatic extraction of abnormalities by learning from experiences. Hence, for PMT, the DBN models have been trained to extract the HD by pairing two different variables: Steering Angle-Power (SP), Velocity-Power (VP) and Steering Angle-Velocity (SV).
\vspace{-2mm}
\subsubsection{Testing phase}
\begin{figure}
\centering
 	(a)\includegraphics[height=2.5cm, width=12cm]{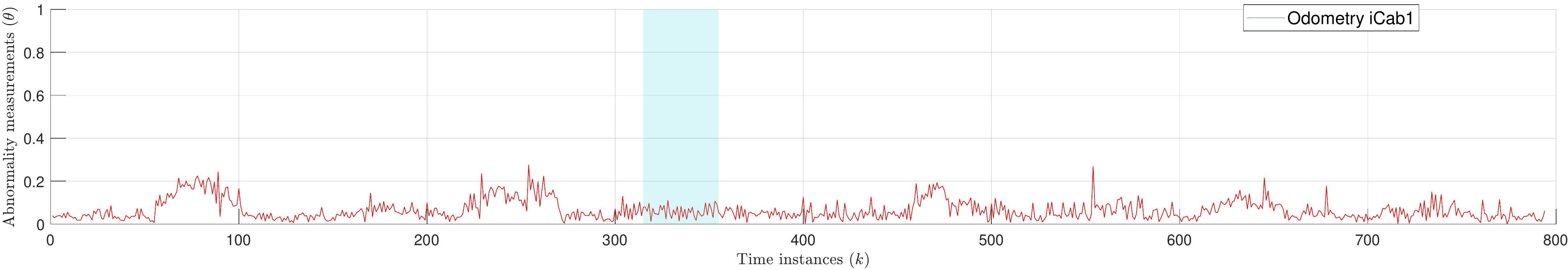}
 	(b)\includegraphics[height=2.5cm, width=12cm]{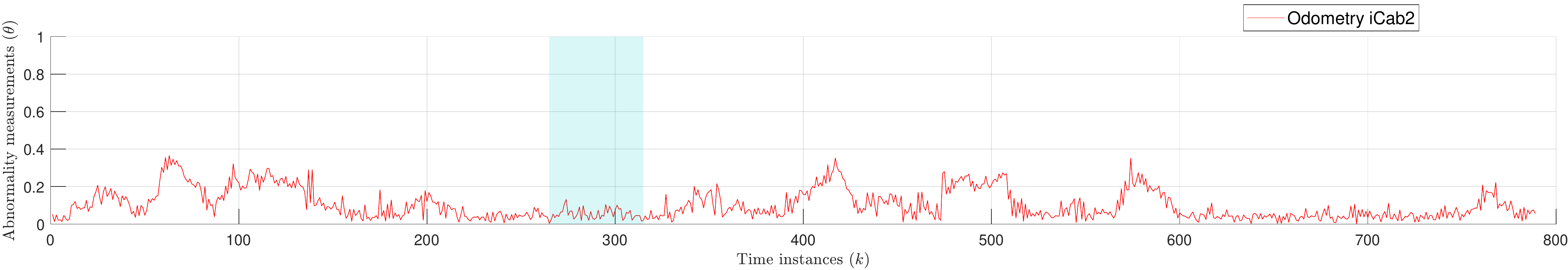}
\caption{Abnormality measurements for odometry: (a) $iCab1$, (b) $iCab2$}
\label{fig:XYcomb}
\centering
    (a)\includegraphics[height=2.5cm, width=12cm]{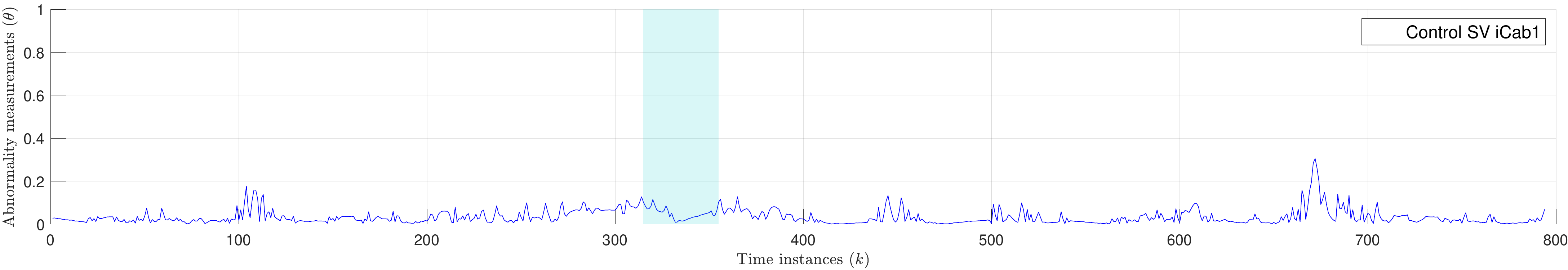}
 	(b)\includegraphics[height=2.5cm, width=12cm]{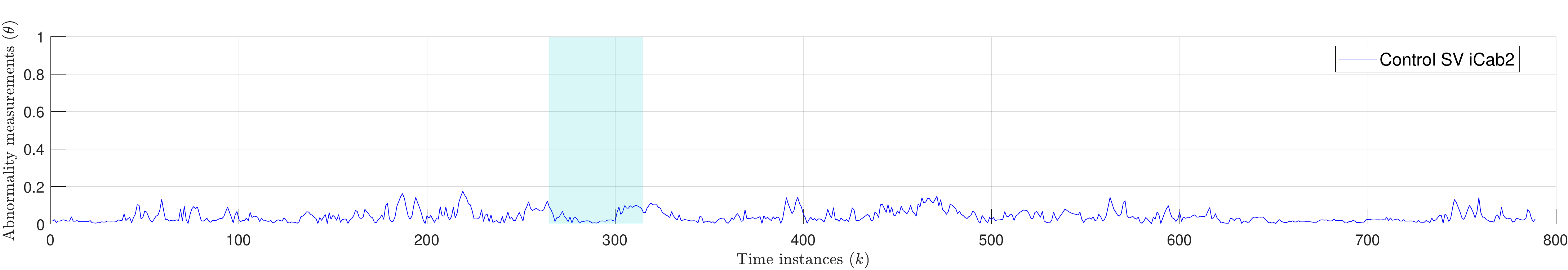}
\caption{Abnormality measurements for control (SV): (a) $iCab1$, (b) $iCab2$}
\label{fig:SVcomb}
\centering
    (a)\includegraphics[width = 1 \linewidth ]{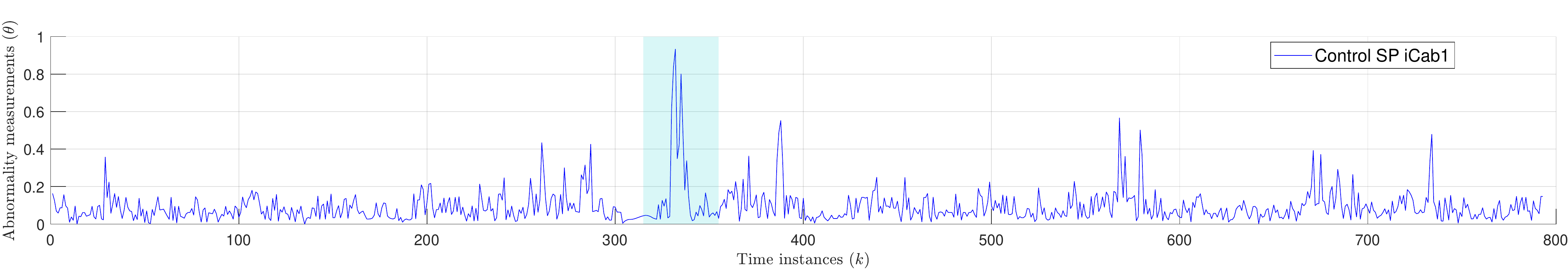}
 	(b)\includegraphics[width = 1 \linewidth ]{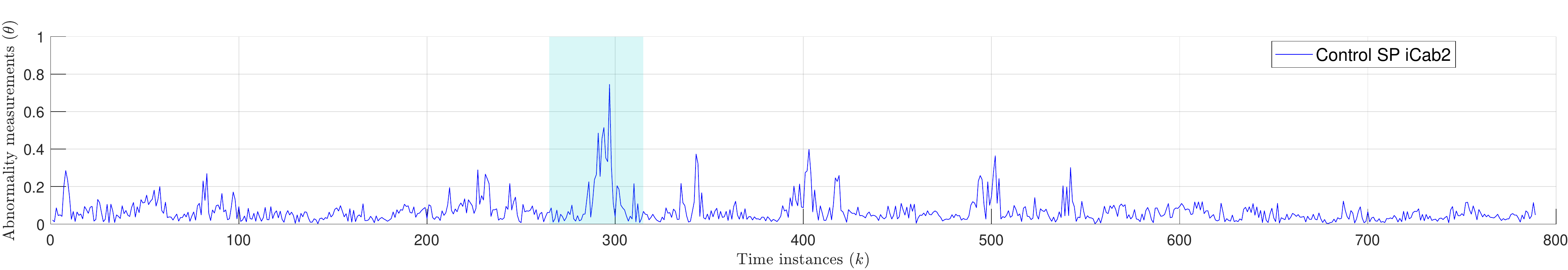}
\caption{Abnormality measurements for control (SP): (a) $iCab1$, (b) $iCab2$}
\label{fig:SPcomb}
\centering
    (a)\includegraphics[width = 1 \linewidth ]{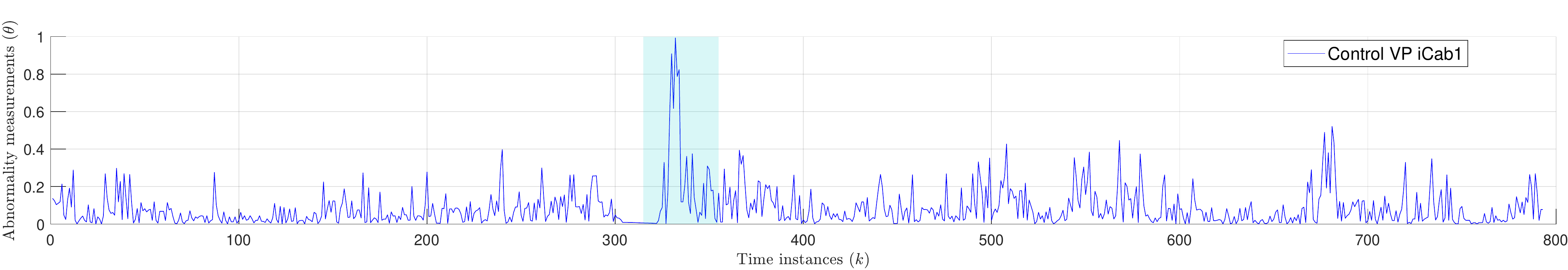}
 	(b)\includegraphics[width = 1 \linewidth ]{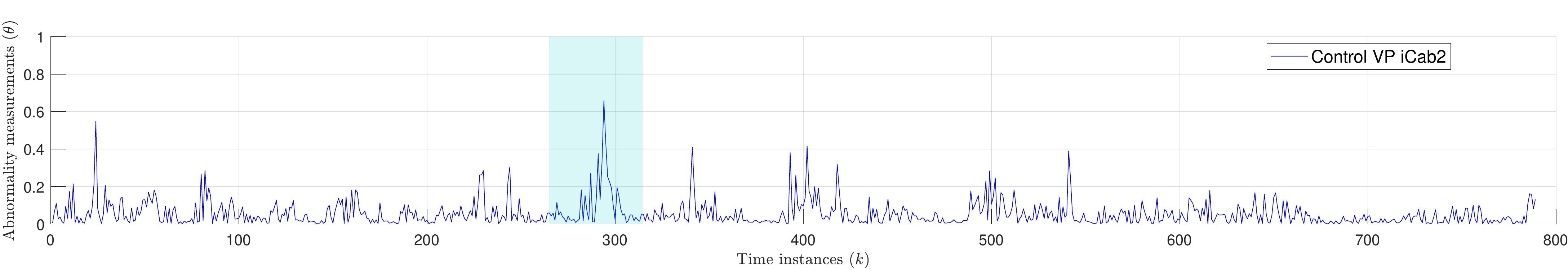}
\caption{Abnormality measurements for control (VP): (a) $iCab1$, (b) $iCab2$}
\label{fig:VPcomb}
\end{figure}

The switching DBN model in this work is designed for the control part of the vehicles. However, we have considered odometry data and tested the performance of the learned DBN. Fig. \ref{fig:XYcomb} shows the plots of abnormality measures by considering odometry data for the vehicle leader (iCab1) and the vehicle follower (iCab2), respectively. During the interval (cyan shaded area) while pedestrian passes and vehicle stops, there isn\textquotesingle’t any significant difference in HD value for iCab1 (leader) as well as iCab2 (follower). This behaviour is due to the fact that, during that interval the vehicles always inside the normal trajectory range. 
However, there are specific intervals when the vehicle deviates from the normal trajectory range, and the HD measures provided a high value of about 0.2 during those intervals. So the learned DBN model was able to predict if any trajectory deviation occurred.
\begin{itemize}
  \item \textbf{Steering Angle-Velocity (SV)}: It is necessary to check that the HD is working in both directions when the metrics involved reflect abnormal behaviour such as power and velocity and in cases when the metric used does not notice when there is an abnormal behavior such as steering angle. For this reason, the pair (S-V) shown in fig.~\ref{fig:SVcomb} displays that this pair is not detecting abnormalities when a pedestrian crosses in front of the vehicle (cyan shaded), which is expected.
 \item \textbf{Steering Angle-Power (SP)}: Fig.~\ref{fig:SPcomb} shows that the HD is high when a pedestrian is crossing in front of the leader vehicle (iCab1). This high value is considered as an abnormality in the behaviour of the leader, and as it is expected, the follower (iCab2) also gives an abnormal behaviour for the platooning task. However, the HD measures for the follower is not as significant as the leader, because the follower was not doing emergence break rather reducing its speed until reaching the minimum distance with the leader.
  \item \textbf{Velocity-Power (VP)}: The last pair tested is velocity and the power consumption which are highly related. In Fig.~\ref{fig:VPcomb}, it is shown the moment when a pedestrian cross in front of the leader in cyan colour, which match with the highest value of the HD. For the follower vehicle, the abnormality measurement is very significant due to the new performance of the vehicle against the emergency brake of the leader. The consecutive peaks in the HD are caused by the high acceleration when the leader vehicle starts moving, but the current distance between them is still lower than the desired. Next peak in HD is caused by the acceleration of the leader and the emergency stop in the follower due to the pedestrian.
 To summarize, the switching DBN learned from SP and VP data were able to predict the unusual situations present; however, odometry and SV combination of control was not showing good combination to detect abnormal behavior.
\end{itemize}

\section{Conclusion and future work} 
\label{section V1}
\vspace{1mm}
It has been proved that the HD for automatically detecting abnormalities in a DBN model learned from experiences, is a possible and plausible solution. The main idea of the proposed method has been demonstrated with the training and testing phase, and the results support that the methodology applied is more useful instead of checking and delimiting each metric of the vehicle depending on the event and defining each upper and lower limits in which an abnormal behaviour is considered.\\
The future work of this new approach could be extended by establishing communication between the objects involved in the tasks and develop collective awareness models. Such models can make the mutual prediction of the future states of the objects involved in the task and enrich the contextual awareness where they operate. Another direction could be the development of an optimized model from the different feature combinations that can be used for the future state predictions of the considered entities. Additionally, the classification of detected abnormality by considering different test scenarios and comparing the performance of abnormality detection by using different metric as abnormality measure could be considered.

\section*{Acknowledgement}
Supported by SEGVAUTO 4.0 P2018/EMT-4362) and CICYT projects (TRA2015-63708-R and TRA2016-78886-C3-1-R).
\\

\begingroup
\let\clearpage\relax
\bibliographystyle{splncs03}
\bibliography{references}
\endgroup

\end{document}